# Adaptive Prompt Learning with Negative Textual Semantics and Uncertainty Modeling for Universal Multi-Source Domain Adaptation

Yuxiang Yang[1], Lu Wen[1], Yuanyuan Xu[1], Jiliu Zhou[1], Yan Wang[1,*]

[1]School of Computer Science, Sichuan University, Chengdu, China

*Abstract*—Universal Multi-source Domain Adaptation (UniMDA) transfers knowledge from multiple labeled source domains to an unlabeled target domain under domain shifts (different data distribution) and class shifts (unknown target classes). Existing solutions focus on excavating image features to detect unknown samples, ignoring abundant information contained in textual semantics. In this paper, we propose an Adaptive Prompt learning with Negative textual semantics and uncErtainty modeling method based on Contrastive Language-Image Pre-training (APNE-CLIP) for UniMDA classification tasks. Concretely, we utilize the CLIP with adaptive prompts to leverage textual information of class semantics and domain representations, helping the model identify unknown samples and address domain shifts. Additionally, we design a novel global instance-level alignment objective by utilizing negative textual semantics to achieve more precise image-text pair alignment. Furthermore, we propose an energy-based uncertainty modeling strategy to enlarge the margin distance between known and unknown samples. Extensive experiments demonstrate the superiority of our proposed method.

## 1. Introduction

Deep neural networks (DNNs) have made significant strides in image classification tasks [1], [2], [3], [4], [5], [6]. However, their performance relies on abundant labeled data, which often drops sharply to a new unlabeled domain. Annotating data in new domains is extremely time-consuming and the inherent distribution (domain) shift [7] between the labeled and unlabeled domains leads to a notable obstacle to model adaptation. To alleviate this, Single-source Domain Adaptation (SDA) methods [8], [9] have been proposed to transfer knowledge in a labeled source domain to an unlabeled target domain. However, in practice, the labeled data can be collected from multiple sources. So Multi-source Domain Adaptation (MDA) [10], [11], [12] is then motivated to deal with the domain shift and transfer knowledge from multiple labeled source domains to an unlabeled target domain via adversarial learning [10] or feature alignment [11]. Nevertheless, these methods all assume that source and target domains share identical classes, which always fails in open-

This work is supported by National Natural Science Foundation of China (NSFC 62371325, 62071314), Sichuan Science and Technology Program 2023YFG0263, 2023YFG0025, 2023YFG0101.

*Corresponding author: Yan Wang (wangyanscu@hotmail.com).

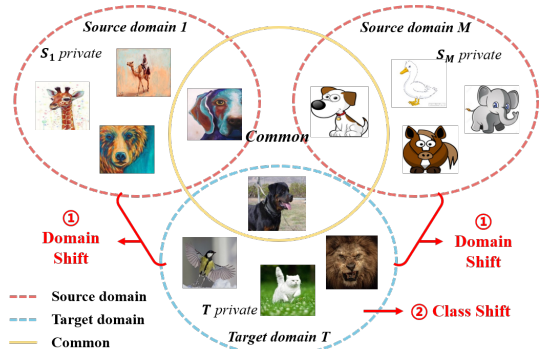

Fig. 1. An illustration to the problem setting of UniMDA where exists both domain shifts (different data distribution) and class shifts (unknown target classes).

world scenarios where different domains may have different class label spaces (i.e., class shift) [13], [14].

To tackle these problems, Universal Multi-source Domain Adaptation (UniMDA) [14] is explored to allow source and target domains to have their private classes. As shown in Fig. 1, UniMDA faces two main challenges. One is how to alleviate domain shifts across multiple domains to extract discriminative representations from common classes. The other is how to address class shifts by detecting the unknown target samples without label information. Until now, there have been only a few studies along the UniMDA research direction. Particularly, [15] utilized a multi-stage learning technique with contrastive learning to predict target labels and filter unknown classes. To measure the class reliability, [14] integrated a pseudo-margin vector into the adversarial training and gained satisfactory performance. However, these methods always focus on excavating image features to boost the UniMDA performance, ignoring the abundant knowledge contained in the textual semantics. Recently, contrastive language-image pre-training (CLIP) models [2], [16] with textual prompts have shown notable efficacy in open-world scenarios, indicating a promising direction for UniMDA.

In this paper, we propose a novel Adaptive Prompt learning with Negative textual semantics and uncErtainty modeling method based on Contrastive Language-Image Pre-training (APNE-CLIP) to better fulfill the UniMDA classification tasks. Our key motivation is to utilize the learnable prompts to enable CLIP to distinguish unknown target samples and address domain shifts simultaneously. Concretely, the learnable prompts adaptively acquire distinct class semantics and domain representations from multiple domains, thus helping CLIP detect unknown target samples and alleviate domain shift effectively. Additionally, considering the problem that some image-text pairs may lack precise one-to-one correspondence, we design a novel instance-level alignment objective by utilizing the textual negative semantics to strengthen image-text pair alignment, thus achieving more precise classification. Furthermore, we propose an energy-based uncertainty modeling strategy. This strategy allocates higher uncertainty energy scores to samples diverging from the known ones, which enlarges the margin distance between known and unknown samples, thus facilitating the accurate detection of unknown samples.

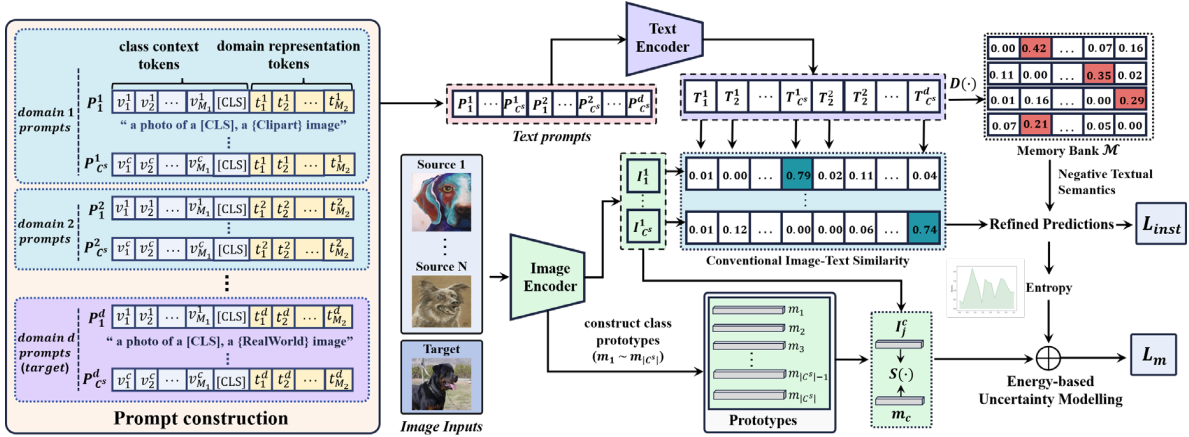

Fig. 2. Overview of the proposed APNE-CLIP. The learnable prompts extract both class semantics and domain representations from multiple domains adaptively. The text prompts and image inputs are respectively fed into the text encoder and image encoder of CLIP.

Our main contributions are listed as follows: (1) We propose a novel APNE-CLIP method for the UniMDA classification tasks to simultaneously alleviate domain and class shifts. (2) We design a novel instance-level alignment objective to leverage textual negative semantics to enhance the image-text alignment, helping the model gain more precise classification. (3) We propose an energy-based uncertainty modeling strategy to expand the margin distance between known and unknown samples, facilitating the model to identify unknown samples. (4) Extensive experiments on three image classification benchmarks have validated the superiority of our APNE-CLIP.

## 2. RELATED WORKS

**Multi-Source Domain Adaptation.** Multi-Source Domain Adaptation (MDA) deals with a practical problem setting where labeled training samples are collected from multiple sources (organizations, devices, sensors, etc.) [10], [12]. Existing MDA methods are under the closed-set assumption where domains share identical classes, which always fails in open-world scenarios where domains may have class shifts [13], [14]. To relieve this issue, Universal Single-Source Domain Adaptation (UniDA) methods have been proposed to use energy discrepancy scores [17], [18] or prediction confidence [13], [19] to identify unknown samples. However, UniDA methods only deal with one single source and overlook the more practical Universal Multi-Source Domain Adaptation (UniMDA) application. Due to the challenging domain and class shifts, only two MDA studies along with UniMDA. Concretely, HyMOS [15] utilized multi-stage learning with contrastive learning for prediction and filtering unknown classes. Besides, UMAN [14] proposed a pseudo-margin vector to estimate the reliability of each known class. Nevertheless, these methods all ignore the critical knowledge in the textual semantics to better address UniMDA tasks.

**Prompt learning.** Prompt learning is a hotspot research that uses appending instructions to the input and pre-training the language model to improve downstream task performance. Specifically, [20] changed manual prompts to learnable ones for better text-image alignment. In SDA, [8] and [9] introduce domain knowledge into prompts to better learn domain-invariant knowledge. Unlike these methods utilize simple prompts or traditional image-text alignment manner, we leverage adaptive prompts and exploit negative textual semantics in CLIP to better facilitate CLIP's capacity.

## 3. METHODOLOGY

### 3.1 Preliminaries and Model Overview

For the problem setting of UniMDA, we follow the descriptions in [14] where there are $N$ labeled source domains $\{S_n\}_{n=1}^N$ and one unlabeled target domain $T$. $S_n$ represents the $n$-th source domain containing a labeled dataset $\mathcal{D}^{S_n} = \{(x_i^{S_n}, y_i^{S_n})\}_{i=1}^{|S_n|}$, where $x_i^{S_n}$ and $y_i^{S_n}$ represent the $i$-th source image and its label, respectively. The target domain $T$ comprises one unlabeled dataset $\mathcal{D}^T = \{x_i^t\}_{i=1}^{|T|}$. For label setting, $C^{S_n}$ denotes the label set of $S_n$ and $C^t$ denotes the target label set. $C^n = C^{S_n} \cap C^t$ denotes the common label set in $S_n$ and $T$, while $\overline{C^{S_n}} = C^{S_n} \setminus C^n$ and $\overline{C^t} = C^t \setminus \{\cup_n C^{S_n}\}$ denotes the private label set of $S_n$ and $T$ respectively. $C = \cup_n C^n$ is the union of common label sets, while $C^s = \cup_n C^{S_n}$ is the union of known source label sets, and $|C^s|$ denotes the total number of known classes. The task for UniMDA is to mitigate domain shifts between $\mathcal{D}^{S_n}$ and $\mathcal{D}^t$ in $C$ and identify target samples to one of the known classes in $C^s$ or the unknown class.

The overview of our APNE-CLIP is depicted in Fig. 2 which comprises two components: a ResNet-based Image Encoder $I$ for generating image embedding, and a Transformer-based Text Encoder $T$ for producing text prompt embedding.

### 3.2 Prompts Construction

In CLIP, prompts typically follow the format "a photo of [CLS]", with [CLS] serving as a placeholder class token that can be replaced with a specific class name. Rather than only employing the manually crafted prompts in the previous works, we train adaptive prompts [9], [20] embedded by the text encoder. For UniMDA, our design includes: 1) class context vectors: $v_i^c$, $i \in [1, M_1]$, $c \in [1, |C^s|]$ to capture more fine-grained representation than only [CLS] tokens, enabling our model to better extract the semantic information of classes. 2) domain representation vectors: $t_j^d$, $j \in [1, M_2]$ where $d=t/s_n$ indicates this image comes from the target or $n$-th source domain and $|d|$ represents the total number of samples in this domain. $M_1$ and $M_2$ represents the number of tokens. Notably, vectors $t_j^d$ are shared across all classes and are proficient in learning domain-specific representations from multiple domains adaptively to alleviate the domain shift. 3) To better exploit the semantic information, we also incorporate the manually crafted prompts into the prompt construction. Notably, as the known classes may include the source private classes, we incorporate the direct "not" information, into the prompt construction. Thus, these prompts can be articulated as either "a photo of $\{c\}$, a $\{d\}$ image" or "a photo of $\{c\}$, not a $\{d\}$ image". So, each learnable prompt $p_c^d$ can be formulated as below:

$$p_c^d = [v_1^c, \ldots, v_{M_1}^c, [CLS_c^d], t_1^d, \ldots, t_{M_2}^d]. \quad (1)$$

*3.3 Novel Instance-Level Alignment*

The optimization of prompts can be divided into two parts: (1) For a labeled source image $x_i^{s_n}$ and its label $y_i^{s_n}$, we optimize the prompts by aligning the outputs from image and text encoders. (2) For an unlabeled target image $x_i^t$, we use CLIP's zero-shot inference ability to generate pseudo-labels $y_i^t$ for image-text alignment, only if the prediction probability exceeds a fixed threshold $\tau$ [9], [20]. Then, the conventional instance-level alignment can be formulated as below:

$$P(y = c) = \frac{exp(<T(p_c^d), I(x_i^d)>/\sigma)}{\Sigma_d \Sigma_{j=1}^{|C^s|} exp(<T(p_j^d), I(x_i^d)>/\sigma)}, \quad (2)$$

where $P(y = c | x_i^d; p_c^d)$ presents the probability of an image sample $x_i^d$ belonging to a specific class, $<,>$ denotes the cosine similarity, and $\sigma$ is a temperature parameter [15], then the instance loss of image-text pairs can be calculated by:

$$\mathcal{L}_{inst} = min_{p_c^d} - \frac{1}{|d|} \Sigma_{i=1}^{|d|} logP(y = y_i^d | x_i^d; p_c^d). \quad (3)$$

By minimizing $\mathcal{L}_{inst}$, the designed learnable prompts can adaptively learn semantic discriminative class information and domain representations from multiple domains, thus facilitating CLIP to detect unknown samples and alleviate the domain shift. However, there is a notable limitation to the conventional instance-level alignment in Eq. (2).

Specifically, for the image embedding of a given sample, Eq. (2) endeavors to ensure the similarity to its corresponding class textual embedding to significantly exceed the similarity with non-related classes. Nevertheless, the conventional instance loss treats all incorrect classes uniformly, disregarding the deep exploration of fine-grained semantic relationships between class textual embeddings (e.g. negative textual semantics) [21], failing to gain a better classification accuracy. In light of this, we design a novel instance-level alignment objective to sufficiently utilize negative textual semantics to accomplish more precise image-text pair alignment. Concretely, we employ a distance function with a memory bank $\mathcal{M}$ to measure and store these negative relations. Each row of $\mathcal{M}$ records the negative distances between one textual class and the others, e.g., $\mathcal{M}_{i,j}$ records the distance between the $i$-th textual class and the $j$-th textual class. The distance function is as follows:

$$D\left(T(p_j^d), T(p_j^d)\right) = 1 - sim\left(g(p_i^d), g(p_j^d)\right). \quad (4)$$

As Eq. (4) illustrates, when the similarity between the embeddings of two text classes is low, the negative semantic distance (relations) should be larger, and they should be distinctly separated within the feature space. Then, we leverage these negative textual semantics to improve the instance-level alignment as below:

$$P(y = c) = \frac{exp(sim(T(p_c^d), I(x_i^d))/\sigma)}{\Sigma_d \Sigma_{j=1}^{|C^s|} exp((sim(T(p_j^d), I(x_i^d)) + \lambda \cdot \mathcal{M}_{j,c})/\sigma)}. \quad (5)$$

The term $\lambda \cdot \mathcal{M}_{j,c}$ can offer additional negative textual semantics between the $j$-th and the $c$-th textual class. By incorporating this strategic instance-level alignment into Eq. (3), we can enhance the alignment between images and their respective textual class and bolster the classification performance of the model.

*3.4 Energy-based uncertainty modeling strategy*

Prior UniDA approaches often use energy discrepancy scores [17], [18] or prediction confidence [13], [19] to identify unknown samples. Nonetheless, these methods, dependent on probability estimates from linear classifiers, can be unreliable and inaccurate, causing suboptimal results. To detect the unknown samples more precisely, we propose an energy-based uncertainty modeling strategy to enlarge the margin distance between known and unknown samples. Concretely, this strategy uses a function $\Omega(\cdot)$ to dynamically assign scores to samples by leveraging their feature-wise similarities with known source class prototypes and prediction probabilities. To this end, we compute the prototype [22] $m_c$ for each of the $|C^s|$ classes by averaging the image embeddings over known source samples as below:

$$m_c = \frac{1}{n_c} \Sigma_{i=1}^N \Sigma_{j=1}^{|s_i|} I(x_j^{s_i}; y_j^{s_i} = c), \quad (6)$$

where $n_c$ represents the total number of samples for a class $c$. For a sample $x_i^d$, we obtain the feature-wise similarity $\phi_c^{x_i^d}$ by measuring the distance between its image embedding and prototype:

$$\phi_c^{x_i^d} = -S(I(x_i^d, m_c)), \quad (7)$$

where $S(\cdot)$ is a similarity function (Euclidean distance). Then we integrate it with the prediction probabilities to define the score $S^{x_i^d}$ of a sample through the dynamic scoring function:

$$S^{x_i^d} = \Omega(x_i^d) = -log\Sigma_{c=1}^{|C^s|} e^{(\phi_c^{x_i^d} + P^{x_i^d})}. \quad (8)$$

This function models the energy-based uncertainty of a sample in feature-wise prototype similarities and class-wise entropy uncertainty, thus providing a more comprehensive uncertainty estimation. Finally, we incorporate these scores into a margin loss:

$$\mathcal{L}_m = \Sigma_{i=1}^N \Sigma_{j=1}^{|s_i|} max\left(0, S^{x_j^{s_i}} - M_s\right) + \Sigma_{i=1}^{|t|} max\left(0, S^{x_i^t} + M_s\right), (9)$$

where $M_s$ is an explicitly defined margin hyper-parameter used to indicate the separation of the known and unknown sample sets when calculating the distance between them. Notably, we only incorporate target samples with prediction probabilities below the threshold $\tau$, as they are more possibly unknown samples. By optimizing $\mathcal{L}_m$, the method assigns higher scores to unknown samples, effectively distancing them from known ones. This separation fosters a more discriminative feature space, thereby improving the detection of unknown samples.

TABLE I. PERFORMACNE COMPARISON OF H-SCORE (%) ON THREE DATASETS IN THE UNIMDA SETTING.

| Protocols | Methods | Office-Home | | | | | Office-31 | | | | DomainNet | | |
|---|---|---|---|---|---|---|---|---|---|---|---|---|---|
| | | →R | →C | →A | →P | Avg | →A | →D | →W | Avg | →S | →C | Avg |
| Source-combine | CLIP [2] | 45.3 | 40.3 | 48.4 | 44.2 | 44.6 | 50.0 | 46.5 | 55.1 | 50.5 | 40.3 | 46.2 | 43.3 |
| | CMU [20] | 77.7 | 61.0 | 64.8 | 71.9 | 68.9 | 72.4 | 74.7 | 71.8 | 73.0 | 40.5 | 41.5 | 41.0 |
| | UniOT [26] | 33.1 | 38.7 | 37.7 | 32.7 | 35.6 | 41.2 | 37.6 | 38.5 | 39.1 | 32.3 | 36.8 | 36.7 |
| | NCAL [27] | 45.4 | 40.7 | 28.8 | 39.5 | 38.6 | 52.0 | 48.5 | 57.1 | 52.5 | 32.1 | 33.0 | 32.6 |
| Multi-source | TFFN [28] | 68.9 | 57.4 | 58.8 | 64.1 | 62.3 | 68.6 | 71.6 | 73.4 | 71.2 | 32.3 | 36.8 | 36.7 |
| | MOSDANET [25] | 67.1 | 52.1 | 53.7 | 61.5 | 58.6 | 69.2 | 58.8 | 65.4 | 64.5 | 34.4 | 41.4 | 37.9 |
| | HyMOS [15] | 74.2 | 66.6 | 67.5 | 71.4 | 69.9 | 60.1 | 76.1 | 74.8 | 70.3 | 51.6 | 54.7 | 53.2 |
| | UMAN [14] | 84.0 | 68.7 | 70.2 | 74.7 | 74.4 | 80.8 | 72.1 | 73.9 | 75.6 | 50.3 | 52.8 | 51.6 |
| | APNE-CLIP(ours) | **87.2** | **69.5** | **83.2** | **86.4** | **81.6** | **84.2** | **76.5** | **76.1** | **78.9** | **58.5** | **62.9** | **60.7** |

TABLE II. PERFORMACNE COMPARISON OF H-SCORE (%) ON THREE DATASETS IN THE OMDA SETTING.

| Protocols | Methods | Office-Home | | | | | Office-31 | | | | DomainNet | | |
|---|---|---|---|---|---|---|---|---|---|---|---|---|---|
| | | →R | →C | →A | →P | Avg | →A | →D | →W | Avg | →S | →C | Avg |
| Source-combine | CLIP [2] | 44.3 | 42.5 | 47.4 | 46.3 | 45.1 | 58.0 | 38.5 | 59.7 | 52.1 | 48.5 | 53.3 | 50.9 |
| | CMU [20] | 70.8 | 50.0 | 58.1 | 69.3 | 62.1 | 56.4 | 64.0 | 61.4 | 60.6 | 38.1 | 35.5 | 36.8 |
| | UniOT [26] | 32.1 | 37.5 | 31.7 | 34.2 | 33.9 | 44.2 | 48.9 | 53.3 | 48.8 | 30.0 | 37.6 | 33.8 |
| | NCAL [27] | 41.0 | 32.5 | 33.8 | 42.2 | 37.4 | 58.0 | 60.1 | 62.7 | 60.3 | 33.6 | 35.4 | 34.5 |
| Multi-source | TFFN [28] | 68.0 | 55.1 | 54.6 | 66.7 | 61.1 | 60.3 | 72.6 | 71.2 | 68.0 | 35.7 | 40.4 | 38.1 |
| | MOSDANET [25] | 65.0 | 51.1 | 54.3 | 65.9 | 59.1 | **73.9** | 71.5 | 60.5 | 68.6 | 40.0 | 39.3 | 39.7 |
| | HyMOS [15] | 71.0 | 64.6 | 62.2 | 71.1 | 67.2 | 60.8 | **89.9** | **90.2** | 80.3 | 57.5 | 61.0 | 59.3 |
| | UMAN [14] | 72.5 | 62.4 | 60.0 | 70.4 | 66.3 | 73.5 | 82.7 | 81.9 | 79.4 | 53.7 | 58.5 | 56.1 |
| | APNE-CLIP(ours) | **73.4** | **65.2** | **74.3** | **86.4** | **74.8** | 70.8 | 88.1 | 87.1 | **82.0** | **61.7** | **63.5** | **62.6** |

Consequently, the overall objective function for training the proposed APNE-CLIP can be formulated as follows:

$$\mathcal{L}_{total} = \mathcal{L}_{inst} + \alpha \cdot \mathcal{L}_m, \quad (10)$$

where $\alpha$ is the hyper-parameter to balance the two losses.

## 4. EXPERIMENTS

### 4.1 Dataset and Evaluation

We perform evaluations on three image classification benchmarks, following the experiment setting in [14], [15]. Office-31 [23] Comprises 4,110 images across 31 classes and three domains: Amazon (A), Webcam (W), and DSLR (D). Office-Home [24] has four domains: Artistic (A), Clip-Art (C), Product (P), and Real-World (R) with a total of 15,588 images in 65 classes. DomainNet [10] contains 6 domains, 345 classes, and around 0.6 million images. Following [14], we include Infograph (I), Painting (P), Sketch (S), and Clipart (C), selecting randomly 50 samples per class or using all the images in case of lower cardinality. For the class settings, we conduct evaluations under two major class shift scenarios: UniMDA and Open-set MDA (OMDA). In UniMDA, both source and target domains have private classes, for Office-31 and Office-Home, we aligned with [14]. For DomainNet, the first 100 alphabetically ordered classes are used as $C$, while the next 145 are unknown $\overline{C^t}$. The first and last 50 classes represent $\overline{C^{sn}}$ for two source domains. In OMDA, only the target domain contains unknown classes, we follow [14], [25]. The detailed class settings are shown in the Supplementary Materials. For evaluation, the energy scores from $\Omega(\cdot)$ reflect the probability of a sample being unknown. Suppose the $u_s$ and $\sigma_s$ are the mean and standard deviation values of the energy scores of known source samples, we set the threshold $\delta = u_s - 2 * \sigma_s$ and divide those target samples with energy scores higher than $\delta$ as unknown samples. Besides, to measure the classification performance, we use the commonly used metric $H_{score} = \frac{2*Acc_k*Acc_u}{Acc_k+Acc_u}$ [14] in universal DA methods, which is the harmonic mean of the accuracy of the known classes $Acc_k$ and that of the unknown classes $Acc_u$.

### 4.2 Training Details

For fair comparisons, we adopt ResNet50 as the backbone for all datasets and its weights are from CLIP [2] and kept frozen during training. The batch size is set as 32, and the prompts are trained using the mini-batch SGD optimizer with a learning rate of 5e-4, following the learning schedule in [9]. For hyper-parameters, the token length $M_1$ and $M_2$ are both set to 16 to contain more prompts. Pseudo-label threshold $\tau$ is set to 0.4 for producing reliable labels. We empirically set $\lambda$ in Eq. (5) to 0.03, $M_s$ in Eq. (9) to 8 and $\alpha$ in Eq. (10) to 0.1. Besides, we maintain a fixed random seed over 3 runs and report their mean results.

### 4.3 Comparison Experiments

To verify the effectiveness of our APNE-CLIP, we compare it with previous state-of-the-art methods: (1) CLIP, Zero-shot manually prompting method; (2) UniDA with Source Combine strategy [11]: CMU [20], UniOT [26], and NCAL [27]; (3) closed-set MDA: TFFN [28]; (4) OMDA:

TABLE III
ABLATION STUDIES ON TWO DATASETS.

| Methods | UniMDA | | OMDA | |
|---|---|---|---|---|
| | H-score | Acc | H-score | Acc |
| Baseline | 77.1 | 76.8 | 72.3 | 71.9 |
| + new $\mathcal{L}_{inst}$ | 78.2 | 78.7 | 73.6 | 74.2 |
| + $\mathcal{L}_m$ (Full) | 81.6 | 82.4 | 74.8 | 75.5 |
| w/o $v_i^c$ | 81.0 | 81.9 | 74.2 | 74.8 |
| w/o($v_i^c$ & $t_j^d$) | 80.2 | 80.6 | 73.4 | 74.2 |

TABLE IV. ANALYSIS OF THE CLIP BACKBONE ON OFFICE-HOME IN UNIMDA SEETING.

| Methods | UNIMDA | | | | |
|---|---|---|---|---|---|
| | →R | →C | →A | →P | Avg |
| CLIP [2] | 45.3 | 40.3 | 48.4 | 44.2 | 44.6 |
| HyMOS [15] | 74.2 | 66.6 | 67.5 | 71.4 | 69.9 |
| HyMOS+CLIP* | 74.8 | 65.2 | 69.5 | 72.4 | 70.5 (↑0.6) |
| UMAN [14] | 84.0 | 68.7 | 70.2 | 74.7 | 74.4 |
| UMAN+CLIP* | 83.6 | 69.2 | 71.0 | 75.0 | 74.7 (↑0.3) |
| APNE-CLIP(ours) | **87.2** | **69.5** | **83.2** | **86.4** | **81.6** |

TABLE V. ANALYSIS OF THE CLIP BACKBONE ON OFFICE-HOME IN OMDA SEETING.

| Methods | OMDA | | | | |
|---|---|---|---|---|---|
| | →R | →C | →A | →P | Avg |
| CLIP [2] | 44.3 | 42.5 | 47.4 | 46.3 | 45.1 |
| HyMOS [15] | 71.0 | 64.6 | 62.2 | 71.1 | 67.2 |
| HyMOS+CLIP* | 71.2 | 64.5 | 61.8 | 70.1 | 66.9 (↓0.3) |
| UMAN [14] | 72.5 | 62.4 | 60.0 | 70.4 | 66.3 |
| UMAN+CLIP* | 72.7 | 62.6 | 60.8 | 71.1 | 66.8 (↑0.5) |
| APNE-CLIP(ours) | **73.4** | **65.2** | **74.3** | **86.4** | **74.8** |

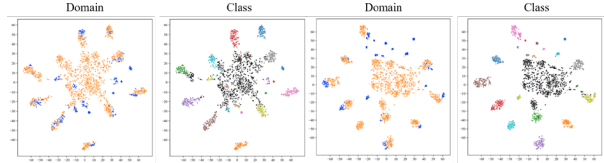

Fig. 3. The left two are t-SNE feature visualizations of UMAN, the rest are ours. For domain, blue represents the source domain and orange refers to the target domain. For class, black plots are unknown samples, others are known samples.

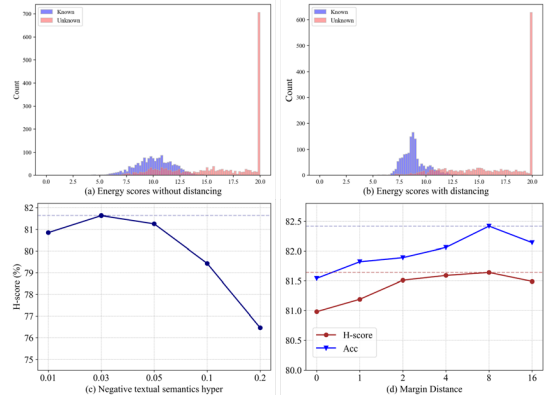

Fig. 4. Analytical study. (a & b) Histogram of energy scores on '→ A ' task in Office-Home of UniMDA setting; (c) Varying the hyper-parameter λ; (d) Varying the margin distance $M_s$.

MOSDANET [25], and HyMOS [14]; (5) UniMDA: UMAN [13]. To ensure a fair comparison, the results of these methods are obtained either from their respective papers or by reimplemented using their released code. As shown in Table I and Table II, we bold the best results. For Office31, APNE-CLIP achieves the best average performance of 78.9% and 82.0% under UniMDA and OMDA settings. On the Office-Home dataset, APNE-CLIP exhibits the highest H-score for all tasks with a significant improvement of 6.8% and 7.6% over the second-best method UMAN and HyMOS in UniMDA and OMDA settings, respectively. On DomainNet, our APNE-CLIP surpasses a huge average gain of 5.5% over the best competitor method HyMOS.

*4.4 Analytical Experiments*

**Feature Visualizations.** We provide t-SNE visualizations of our model and the UMAN on the '→A' task in Office-31. As shown in Fig. 3, compared to UMAN, our method forms more compact clusters with more distinct boundaries. This indicates that APNE-CLIP can achieve the common classes mixing well and separate most target private samples from the common samples, showcasing better transfer capability.

**Ablation Study.** To validate the effects of key components in our model, we progressively conduct ablative experiments on Office-Home in two class settings. The baseline indicates using our adaptive prompts and the conventional instance loss. As observed from Table III: the integration of the new $\mathcal{L}_{inst}$ can yield a huge average improvement of 1.2% in H-score and 2.1% in Acc on two class shift settings, respectively, demonstrating its effectiveness in enhancing image-text pair alignment.

Moreover, without adaptive prompts $v_i^c$ and $t_j^d$ to facilitate the model gaining class and domain representations, our model has an evident performance deterioration, which manifests the effectiveness of adaptive prompts in distinguishing unknown target samples and alleviating the domain shifts, respectively.

**Effect of Energy-based Uncertainty Modeling Strategy.** To show the effectiveness of our energy-based uncertainty modeling strategy, we train the model without $\mathcal{L}_m$ and utilize the threshold δ to identify unknown samples. Table III shows that our $\mathcal{L}_m$ can obtain higher performance in all settings. As the $\mathcal{L}_m$ can maximize the difference between the known and unknown sample sets, thus distancing the distribution gap of energy scores between the known and unknown samples (See Fig.4 (a & b)). Moreover, we assess the impact of the margin distance $M_s$ on H-score and Acc performance. As demonstrated in Fig. 4 (c), we find a relatively large margin distance typically can enhance model performance, and when $M_s$ is set to 8, the model obtains the best performance.

**Analysis of the CLIP Backbone.** To analyze whether the performance gain of APNE-CLIP is simply brought by CLIP's strong backbone, we swap the backbones of HyMOS and UMAN to CLIP's image encoder and utilize a simple prompt learning strategy [2]. Table IV and Table V ( * denotes our implementations) show that simply using textual information is not always effective and even can bring the opposite effect (0.3% drop of HyMOS in OMDA). Overall, while CLIP can provide a strong backbone, however, without appropriate design, it is not universally superior in other settings [29], [30]. As observed in Table I and Table II, with our novel design to mitigate domain shifts and detect unknown classes, our method can consistently outperform CLIP in all settings on all datasets.

**Parameter Sensitivity.** We also conduct parameter sensitivity tests on Office-Home for the negative textual semantics hyper-parameter λ. We select values from {0.01, 0.03, 0.05, 0.1, 0.2}. As seen from the experimental results in Fig. 4 (d), the performance gains improvements with a small value of λ as over-tuning may undermine the semantic information of the text. Ultimately, we choose λ =0.03. More qualitative experiments, including *the identification of unknown classes*, are given in the Supplementary Materials.

## 5. CONCLUSIONS

In this paper, we propose the APNE-CLIP to tackle the UniMDA classification tasks. By incorporating adaptive prompts, we equip CLIP to gain both class semantics and domain representations, thus promoting its proficiency in identifying unknown samples and alleviating domain shifts. Additionally, we design a novel global instance-level alignment objective to utilize negative textual semantics for enhanced image-text alignment. Furthermore, we propose an energy-based uncertainty modeling strategy to increase the margin distance between known and unknown samples, thereby enhancing the accurate detection of unknown samples. Extensive experiments on three challenging datasets substantiate the effectiveness of our proposed APNE-CLIP.

## 6. REFERENCES

[1] K. He, X. Zhang, S. Ren, and J. Sun, "Deep Residual Learning for Image Recognition," in 2016 IEEE Conference on Computer Vision and Pattern Recognition (CVPR), Las Vegas, NV, USA, Jun. 2016. doi: 10.1109/cvpr.2016.90.

[2] A. Radford et al., "Learning Transferable Visual Models From Natural Language Supervision," International Conference on Machine Learning, International Conference on Machine Learning, Jul. 2021.

[3] Z. Du et al., "LION: Label Disambiguation for Semi-supervised Facial Expression Recognition with Progressive Negative Learning," in Proceedings of the Thirty-Second International Joint Conference on Artificial Intelligence, Macau, SAR China, Aug. 2023. doi: 10.24963/ijcai.2023/78.

[4] Y. Yang, Y. Hou, L. Wen, P. Zeng, and Y. Wang, "Semantic-aware Adaptive Prompt Learning for Universal Multi-source Domain Adaptation," IEEE Signal Processing Letters, 2024.

[5] Q. Ma, J. Gao, B. Zhan, Y. Guo, J. Zhou, and Y. Wang, "Rethinking Safe Semi-supervised Learning: Transferring the Open-set Problem to A Close-set One," in Proceedings of the IEEE/CVF International Conference on Computer Vision, 2023, pp. 16370-16379.

[6] Y. Shi et al., "ASMFS: Adaptive-similarity-based multi-modality feature selection for classification of Alzheimer's disease," Pattern Recognition, p. 108566, Jun. 2022, doi: 10.1016/j.patcog.2022.108566.

[7] B. Gong, K. Grauman, and F. Sha, "Connecting the Dots with Landmarks: Discriminatively Learning Domain-Invariant Features for Unsupervised Domain Adaptation," International Conference on Machine Learning, International Conference on Machine Learning, Jun. 2013.

[8] S. Bai et al., "Prompt-Based Distribution Alignment for Unsupervised Domain Adaptation," Proceedings of the AAAI Conference on Artificial Intelligence, vol. 38, no. 2, pp. 729–737, Mar. 2024, doi: 10.1609/aaai.v38i2.27830.

[9] C. Ge et al., "Domain Adaptation via Prompt Learning," IEEE Transactions on Neural Networks and Learning Systems, pp. 1–11, Jan. 2024, doi: 10.1109/tnnls.2023.3327962.

[10] Z. Cai, T. Zhang, and X.-Y. Jing, "Dual Re-Weighting Network for Multi-Source Domain Adaptation," in 2022 IEEE International Conference on Multimedia and Expo (ICME), Taipei, Taiwan, Jul. 2022. doi: 10.1109/icme52920.2022.9859615.

[11] X. Peng, Q. Bai, X. Xia, Z. Huang, K. Saenko, and B. Wang, "Moment Matching for Multi-Source Domain Adaptation," in 2019 IEEE/CVF International Conference on Computer Vision (ICCV), Seoul, Korea (South), Oct. 2019. doi: 10.1109/iccv.2019.00149.

[12] Y. Yang, L. Wen, P. Zeng, B. Yan, and Y. Wang, "DANE: A Dual-level Alignment Network with Ensemble Learning for Multi-Source Domain Adaptation," IEEE Transactions on Instrumentation and Measurement, 2024.

[13] K. You, M. Long, Z. Cao, J. Wang, and M. I. Jordan, "Universal Domain Adaptation," in 2019 IEEE/CVF Conference on Computer Vision and Pattern Recognition (CVPR), Long Beach, CA, USA, Jun. 2019. doi: 10.1109/cvpr.2019.00283.

[14] Y. Yin, Z. Yang, H. Hu, and X. Wu, "Universal multi-Source domain adaptation for image classification," Pattern Recognition, vol. 121, p. 108238, Jan. 2022, doi: 10.1016/j.patcog.2021.108238.

[15] S. Bucci, F. Borlino, B. Caputo, and T. Tommasi, "Distance-based Hyperspherical Classification for Multi-source Open-Set Domain Adaptation," in 2022 IEEE/CVF Winter Conference on Applications of Computer Vision (WACV), Waikoloa, HI, USA, Jan. 2022. doi: 10.1109/wacv51458.2022.00110.

[16] H. Wang, Y. Li, H. Yao, and X. Li, "CLIPN for Zero-Shot OOD Detection: Teaching CLIP to Say No," in 2023 IEEE/CVF International Conference on Computer Vision (ICCV), Oct. 2023. doi: 10.1109/iccv51070.2023.00173.

[17] M. Shen, A. J. Ma, and P. C. Yuen, "E 2: Entropy Discrimination and Energy Optimization for Source-free Universal Domain Adaptation," in 2023 IEEE International Conference on Multimedia and Expo (ICME), 2023: IEEE, pp. 2705-2710.

[18] W. Liu, X. Wang, JohnD. Owens, and Y. Li, "Energy-based out-of-distribution detection," in Advances in neural information processing systems, Oct. 2020, vol. 33, no. 21464--21475.

[19] B. Fu, Z. Cao, M. Long, and J. Wang, "Learning to Detect Open Classes for Universal Domain Adaptation," in Computer Vision – ECCV 2020, Lecture Notes in Computer Science, 2020, pp. 567–583. doi: 10.1007/978-3-030-58555-6_34.

[20] K. Zhou, J. Yang, C. Loy, and Z. Liu, "Conditional Prompt Learning for Vision-Language Models," in 2022 IEEE/CVF Conference on Computer Vision and Pattern Recognition (CVPR), Jun. 2022. doi: 10.1109/cvpr52688.2022.01631.

[21] X. Huo, L. Xie, H. Hu, W. Zhou, H. Li, and Q. Tian, "Domain-Agnostic Prior for Transfer Semantic Segmentation," in 2022 IEEE/CVF Conference on Computer Vision and Pattern Recognition (CVPR), Jun. 2022. doi: 10.1109/cvpr52688.2022.00694.

[22] K. Li, C. Liu, H. Zhao, Y. Zhang, and Y. Fu, "ECACL: A Holistic Framework for Semi-Supervised Domain Adaptation," in 2021 IEEE/CVF International Conference on Computer Vision (ICCV), Montreal, QC, Canada, Oct. 2021. doi: 10.1109/iccv48922.2021.00846.

[23] R. Ramakrishnan, B. Nagabandi, J. Eusebio, S. Chakraborty, H. Venkateswara, and S. Panchanathan, "Deep Hashing Network for Unsupervised Domain Adaptation," in Domain Adaptation in Computer Vision with Deep Learning, 2020, pp. 57–74. doi: 10.1007/978-3-030-45529-3_4.

[24] K. Saenko, B. Kulis, M. Fritz, and T. Darrell, "Adapting Visual Category Models to New Domains," in Computer Vision – ECCV 2010, Lecture Notes in Computer Science, 2010, pp. 213–226. doi: 10.1007/978-3-642-15561-1_16.

[25] S. Rakshit, D. Tamboli, P. S. Meshram, B. Banerjee, G. Roig, and S. Chaudhuri, "Multi-source Open-Set Deep Adversarial Domain Adaptation," in Computer Vision – ECCV 2020, Lecture Notes in Computer Science, 2020, pp. 735–750. doi: 10.1007/978-3-030-58574-7_44.

[26] W. Chang, Y. Shi, H. Tuan, and J. Wang, "Unified optimal transport framework for universal domain adaptation," Advances in Neural Information Processing Systems, vol. 35, pp. 29512-29524, 2022.

[27] Su, Z. Han, R. He, B. Wei, X. He, and Y. Yin, "Neighborhood-based credibility anchor learning for universal domain adaptation," Pattern Recognition, vol. 142, p. 109686, Oct. 2023, doi: 10.1016/j.patcog.2023.109686.

[28] Y. Li, S. Wang, B. Wang, Z. Hao, and H. Chai, "Transferable feature filtration network for multi-source domain adaptation," Knowledge-Based Systems, vol. 260, p. 110113, Jan. 2023, doi: 10.1016/j.knosys.2022.110113.

[29] B. Devillers, B. Choksi, R. Bielawski, and R. VanRullen, "Does language help generalization in vision models?," in Proceedings of the 25th Conference on Computational Natural Language Learning, Jan. 2021. doi: 10.18653/v1/2021.conll-1.13.

[30] J. Yang et al., "Unified Contrastive Learning in Image-Text-Label Space," in 2022 IEEE/CVF Conference on Computer Vision and Pattern Recognition (CVPR), Jun. 2022. doi: 10.1109/cvpr52688.2022.01857.

[31] B. Gong, Y. Shi, F. Sha, and K. Grauman, "Geodesic flow kernel for unsupervised domain adaptation," in 2012 IEEE conference on computer vision and pattern recognition, 2012: IEEE, pp. 2066-2073.

# 7. SUPPLEMENTARY MATERIAL

## 7.1 Dataset split details

**Office-31** [23]. Following previous works [14], [15], [25], in the UniMDA setting, both source and target domains have private classes. We select the 10 classes shared by Office-31 [23] and Caltech-256 [31] used as common label set $C$, while the rest classes are sorted in alphabetical order. Specifically, the last 10 classes are unknown and used as $\overline{C^t}$, the first 5 classes and the rest are used as $\overline{C^{sn}}$ for the two source domains respectively. In the OMDA setting, only the target domain contains unknown classes, we follow [14], [25], setting the first 20 classes in alphabetic order as known, while the remaining 11 classes are unknown.

**Office-Home** [24]. In the UniMDA setting, we follow [14]. The last 50 classes in alphabetic order are unknown target classes and are used as $\overline{C^t}$, and the remaining first 10 classes are used as $C$, and the last 5 classes are split and used as $\overline{C^{sn}}$ for the three source domains respectively (each source domain has two private classes). For the OMDA setting, we also use the same class split on Office-Home as in previous work [25], [15] and set the first 45 classes in alphabetic order as known classes, and the remaining 20 are unknown.

**DomainNet** [10]. Following previous works [14], [25], In the UniMDA setting: the first 100 classes in alphabetic order are known, and used as $C$, while the remaining 145 are unknown and used as $\overline{C^t}$. The remaining first 50 classes and the last 50 classes are used as $\overline{C^{sn}}$ for the two source domains respectively. In the OMDA setting, the first 100 classes in alphabetic order are known, while the remaining 245 are unknown target classes. Notably, in real-world conditions, it is difficult to have direct knowledge about the number of unknown classes in the unlabeled target domain. Therefore, we introduced a commonness value $\beta = |C^s \cap C^t| / |C^s \cup C^t|$ to show the distance of the label sets among multiple domains.

## 7.2 More Analytical Experiments

**Identification of Unknown Classes**. Since the identification ability to distinguish between known and unknown samples is important in open-world scenarios. We further evaluate the identification ability for unknown classes of our method APNE-CLIP by comparing it with the methods CMU [20], HyMOS [15], and UMAN [14]. Similar to [15], we also use AUC to evaluate the identification ability, in which we regard the unknown target data as a negative class and the others as a positive one. Table VI and Table VII list the comparison results under UniMDA and OMDA settings in the Office-Home dataset. It can be found that the recent HyMOS and UMAN lag behind our method by an average of 6.7% and 5.2% in AUC, respectively, which demonstrates the effectiveness of our method in the identification of unknown classes.

**Varying Target Private Label set $\overline{C^t}$**. To assess the robustness of our method under the different number of unknown classes, we compare the behavior of APNE-CLIP with HyMOS and UMAN with fixed common labels set $C$ and $\beta$, while increasing the number of unknown classes in $\overline{C^t}$. In this analysis, we use the large-scale DomainNet dataset to conduct experiments, The H-score result is shown in Fig. 5 (a). Our APNE-CLIP consistently outperforms HyMOS and UMAN with $\overline{C^t}$ varying, indicating that it is robust to the change of unknown target class numbers.

TABLE VI
AUC (%) FOR UNKNOWN CLASS IDENTIFICATION ON OFFICE-HOME IN UNIMDA SETTING.

| Methods | Office-Home | | | | |
|---|---|---|---|---|---|
| | →R | →C | →A | →P | Avg |
| CMU [20] | 87.2 | 69.6 | 75.2 | 80.9 | 78.2 |
| HyMOS [15] | 86.1 | 72.4 | 78.3 | 80.6 | 79.4 |
| UMAN [14] | 92.5 | 73.8 | 82.6 | 82.0 | 82.7 |
| APNE-CLIP(ours) | **96.9** | **74.5** | **89.4** | **91.3** | **88.0** |

TABLE VII
AUC (%) FOR UNKNOWN CLASS IDENTIFICATION ON OFFICE-HOME IN OMDA SETTING.

| Methods | Office-Home | | | | |
|---|---|---|---|---|---|
| | →R | →C | →A | →P | Avg |
| CMU [20] | 80.6 | 69.2 | 72.9 | 78.4 | 75.3 |
| HyMOS [15] | 81.1 | 76.4 | 75.3 | 79.6 | 78.1 |
| UMAN [14] | 86.6 | 72.1 | 73.4 | 78.6 | 77.7 |
| APNE-CLIP (ours) | **89.7** | **74.8** | **82.2** | **84.5** | **82.8** |

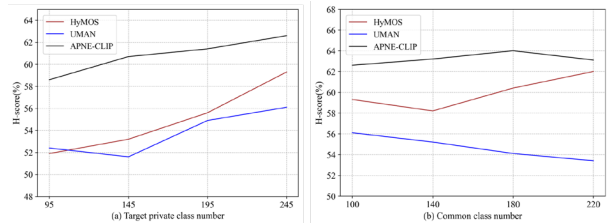

Fig. 5. Various Analytical studies for the number of target private classes and common classes.

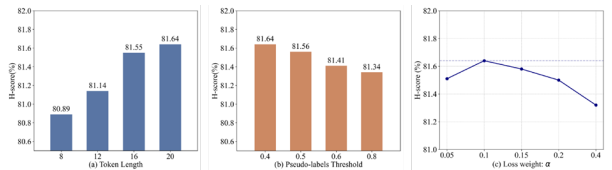

Fig. 6. Parameters sensitivity tests for token length, pseudo-labels threshold and loss weight.

**Varying Common Label $C$**. We also conduct an analysis experiment of APNE-CLIP under different common class numbers in the DomainNet dataset. The number of $\overline{C^t}$ is fixed with $(\cup_n C^{sn}) + 45$ with the number of common classes increasing. Fig. 5 (b) indicates that HyMOS and UMAN are more sensitive to the size of the common label, while APNE-CLIP is not, further demonstrating the effectiveness and robustness of our APNE-CLIP.

**Analysis of Prompts Design.** For prompts design, we conduct hyperparameters sensitivity tests on the Office-Home dataset for the token length and the pseudo-labels threshold. We set for $M_1 = M_2$ for simplification and select the values of them from {8, 12, 16, 20}, the threshold $\tau$ from {0.4, 0.5, 0.6, 0.8}. As seen from the experimental results in Fig. 6 (a & b), the performance gains improvements with longer prompt token length, and our model is not sensitive to $\tau$ due to the balance between the quality and quantity of target samples. Ultimately, we choose $\tau = 0.4$ and $M_1 = M_2 = 16$ to offer an optimal compromise between the performance and computational efficiency.

**Hyperparameter sensitivity of loss weight $\alpha$.** To show the sensitivity of our APNE-CLIP to the margin loss weight $\alpha$, we conducted control experiments on Office-Home under the UniMDA setting and presented the results in Fig. 6 (c). Within a wide range of $\alpha \in [0.05, 0.4]$, the performance changes very little, showing that APNE-CLIP is robust to the selection of $\alpha$. Finally, we choose $\alpha$=0.1 to gain the optimal performance.